\documentclass[12pt]{spieman}  
\usepackage{amsmath,amsfonts,amssymb}
\usepackage{graphicx}
\usepackage{setspace}
\usepackage{tocloft}
\usepackage{multirow}
\usepackage{comment}
\usepackage{makecell}
\usepackage{caption}
\usepackage{subcaption}
\newcolumntype{P}[1]{>{\centering\arraybackslash}p{#1}}
\usepackage{outlines}
\usepackage{hyperref}
\usepackage{lineno}

\title{Continual Domain Adaptation on Aerial Images under Gradually Degrading Weather}

\author{Chowdhury Sadman Jahan}
\author{Andreas Savakis}
\affil{Rochester Institute of Technology, Rochester, NY 14623, USA}

\cftpagenumbersoff{figure}
\cftpagenumbersoff{table} 
\begin{document} 
\maketitle

\begin{abstract}
Domain adaptation (DA) strives to mitigate the domain gap between the source domain where a model is trained, and the target domain where the model is deployed. 
When a deep learning model is deployed on an aerial platform, 
it may face gradually degrading weather conditions during operation, leading to widening domain gaps between the training data and the encountered evaluation data. 
We synthesize two such gradually worsening weather conditions on real images from two existing aerial imagery datasets, generating a total of four benchmark datasets. 
Under the continual, or test-time adaptation setting, we evaluate three DA models on our datasets: a baseline standard DA model and two continual DA models.
In such setting, the models can access only one small portion, or one batch of the target data at a time, and adaptation takes place continually, and over only one epoch of the data. 
The combination of the constraints of continual adaptation, and gradually deteriorating weather conditions provide the practical DA scenario for aerial deployment. 
Among the evaluated models, we consider both convolutional and transformer architectures for comparison. 
We discover stability issues during adaptation for existing buffer-fed continual DA methods, and propose gradient normalization as a simple solution to curb training instability. 
\end{abstract}


{\noindent \footnotesize\textbf{*}Andreas Savakis,  \linkable{andreas.savakis@rit.edu} }

\begin{spacing}{2}   

\section{Introduction}

Deep learning methods have made significant progress dealing with aerial images in terms of fundamental computer vision tasks, such as classification \cite{pritt2017satellite,kyrkou2019deep,kussul2017deep}, segmentation \cite{kaiser2017learning,yue2019treeunet,yeung2022deep}, object detection \cite{sommer2017deep,cheng2016survey}, etc. 
Various sensing modalities of aerial imagery have been considered, such as RGB images \cite{kussul2017deep,kaiser2017learning}, synthetic aperture radar images \cite{jahan2022cross, jahanivmsp2022, jahan2023balanced}, and multi-spectral or hyper-spectral images \cite{nalepa2021towards}. 
However, deep learning models 
face the challenge of domain shift when deployed in an environment outside the distribution of the training data. 
Due to the lack of large-scale diverse
annotated datasets of aerial images, 
addressing the domain shift becomes even more pressing in the remote sensing domain.

Domain shift occurs when the class conditional distribution of the evaluation dataset is different than that of the training dataset, even though the two datasets may have the same classes \cite{torralba2011unbiased}.
Due to such distribution shift, deep learning models trained on one distribution do not transfer well to an unseen distribution.
The domain shift or gap can manifest drastically when a deep learning model trained in one domain is evaluated in a significantly different domain, such as the scenario where a model trained on synthetic images is evaluated on measured, real-life images. 
Airborne platforms, such as 
Unmanned Aerial Vehicles (UAVs) and satellites, are more likely to experience a gradual domain shift, for example when the weather conditions become gradually adverse during operation. 
A model trained on clear weather images may find itself in the middle of a snowstorm during flight. 
In this paper, we specifically address such practical, and critical gradual distribution shift problems that deep learning models in UAVs and satellites may encounter.  


Domain adaptation (DA) is the process of mitigating the adverse effects of distribution shift between training and testing datasets. 
Within the scope of domain adaptation, the labels for the training data, or source domain, are assumed to be available, while those of the testing data, or target domain, are unavailable. 
Existing domain adaptation methods may try to align the subspace, or manifold between the source and target images by projecting the features to a new space \cite{chen2020domain,sun2015subspace,minnehan2019deep}, or by adversarially constructing a common domain-invariant representation space \cite{tzeng2017adversarial,zhang2018collaborative}. 
These models require that both the source, and target domain data be available in their entirety at the same time for adaptation. 

However, access to the source domain becomes impractical for edge platforms, such as UAVs and satellites, because of limited data storage space and computational resources. 
To overcome these limitations, we explore source-free domain adaptation \cite{shot, yang2021generalized, ding2022source}, where the source model is first trained separately with the source data and labels, and the source-trained model is then adapted on the target data without accessing any of the source data. 
This alleviates the need to store the source data, thus leaving a much smaller data storage footprint needed for target adaptation. 
Source-free DA still requires access to the entire target dataset during adaptation. 
For practical purposes, adaptation needs to happen continually, or during test-time, where small batches of target domain data become available, one at a time. 
This motivates the need for continual domain adaptation \cite{taufique2022unsupervised}. 

The continual DA framework is illustrated in Figure \ref{fig:continualDA}.
The latest research on deep domain adaptation has addressed this issue by utilizing a memory buffer to selectively store and replay some samples from earlier batches. While the continual model ConDA \cite{taufique2023condaTAI} performed continual DA across significantly different domains, UCL-GV \cite{taufique2022unsupervised} dealt with DA across gradually varying domains. 

\begin{figure}
\centering
\begin{minipage}[b]{0.9\linewidth}
  \centering
  \centerline{\includegraphics[width=\textwidth]{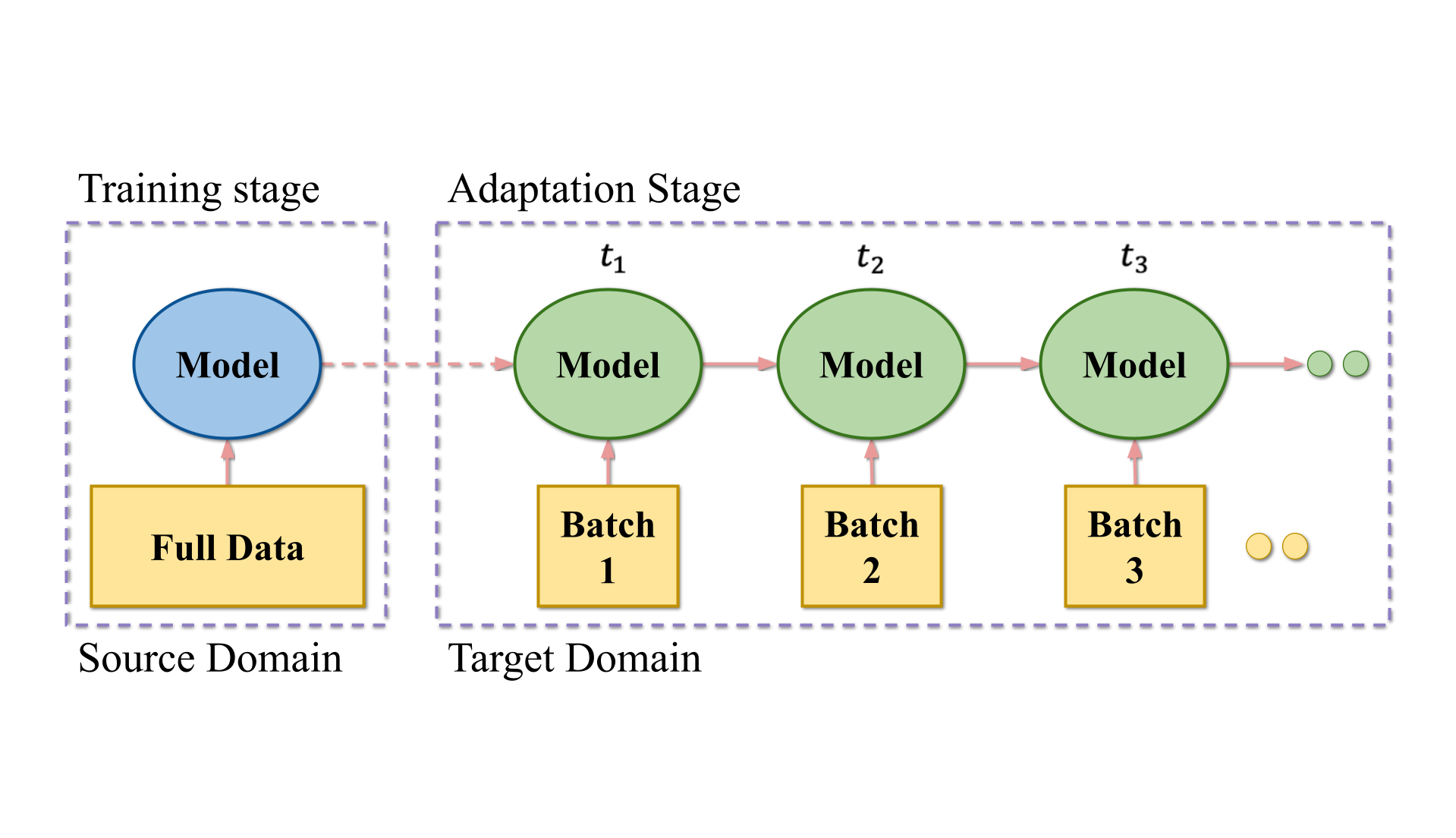}}
  \label{fig:continualDA}
 \end{minipage}
\caption{Illustration of continual DA. The source model is trained with the entire source data and, when deployed, the model is adapted on each incoming target batch, one at a time. Once adapted on one batch, the model is fed the next batch for adaptation with no access to earlier batches.}
\label{fig:continualDA}
\end{figure}

Although DA has been extensively studied for ground-level imagery, very few works have explored DA on aerial images. Nagananda et al. \cite{nagananda2021benchmarking} and Xu et al. \cite{xu2022eyes} evaluated the state-of-the-art standard (non-continual) DA methods on aerial datasets. 
Nagananda \textit{et al.} \cite{nagananda2021benchmarking} 
created three pairs of aerial datasets for DA based on common class labels. 
However, both  works \cite{nagananda2021benchmarking, xu2022eyes} dealt with standard DA settings, with sudden and drastic domain shift between the source and the target domains, and did not consider gradually varying domains.
To the best of our knowledge, continual domain adaptation has not yet been studied within the scope of remote sensing datasets and there are no aerial datasets that could be utilized to assess continual DA on gradually changing environments. 

Our work specifically deals with continual DA on aerial imagery, where the target data distribution is gradually shifting away from that of the source data due to inclement weather. 
In this paper, we present four benchmark datasets for assessing domain adaptation on aerial images, given the domain shifts are gradual. 
We consider degradation types of cloud cover, and snowfall on two widely used aerial image datasets AID \cite{xia2017aid} and UCM \cite{yang2010bag}. 
The descriptions of the datasets, the degradation types, and how they were created are described in Section \ref{sec:dataset}. 
We then evaluate one standard source-free DA model \cite{pmlr-v119-liang20a} and two continual DA models \cite{taufique2023condaTAI,taufique2022unsupervised} on our newly constructed aerial datasets. 
We discover that the continual DA models suffer from significant stability issues, that not only harm optimal adaptation but may potentially result in model collapse. 
We propose the simple solution of normalizing gradients before model optimization to increase adaptation stability, and show empirical results to support our claim.
We also replace the original ResNet-50 \cite{He_2016_CVPR} with attention-based transformer networks Vision Transformer (ViT) \cite{vit} and Swin \cite{swin, swinv2}, and evaluate the models with the state-of-the-art feature extractors to explore the effect of stronger backbone architectures on continual DA.
The main contributions of this paper are as follows.


\begin{enumerate}
  \item We create four new evaluation datasets for domain adaptation under gradually degrading weather, based on two existing aerial image datasets and two types of weather degradations (cloud cover and snowfall).
  \begin{itemize}
    \item The cloud cover degradation type ranges from clear sky to extremely cloudy sky with six levels of gradually increasing cloud cover in between.
    \item The snowfall degradation type ranges from clear sky to heavy snowfall with four gradually worsening intermediate snowfall levels.
    \item The above datasets are publicly available at https://github.com/sadman-jahan/AID-UCM-DegradingWeather
  \end{itemize}
  \item We evaluate two state-of-the-art continual source-free DA models \cite{taufique2023condaTAI, taufique2022unsupervised} and one source-free DA model \cite{shot} on these newly created benchmarks.
  \item We improve stability when training continual DA models\cite{taufique2023condaTAI, taufique2022unsupervised}  by introducing gradient normalization as a simple but effective way to increase adaptation stability for continual DA models with buffer replay.
  \item We explore domain adaptation models in the continual DA setting with attention based transformer backbones\cite{vit, swin}.
\end{enumerate}

\section{Related Works}
\subsection{Unsupervised Domain Adaptation}

Domain gap or shift significantly hurts the performance of a model due to distribution discrepancy between the source and target domains.
Minimizing domain divergence was proposed as a means to mitigate this distribution disparity \cite{long2015learning, tzeng2014deep, zellingercentral}. 
Long et al. \cite{long2015learning}, and Tzeng et al. \cite{tzeng2014deep} proposed to minimize the maximum mean discrepancy (MMD), while Zellinger et al. \cite{zellingercentral} suggested to minimize the central moment discrepancy (CMD) between the source and target distributions. 
Sun et al. \cite{sun2016deep}, in turn, proposed to minimize distribution statistics of the second-order to mitigate domain shift. 
With the advent of generative adversarial learning \cite{goodfellow2014generative}, several later methods proposed to adversarially align the source and target data in the latent feature space \cite{ganin2016domain, long2017deep, pei2018multi}. 
Ganin et al. \cite{ganin2016domain} introduced the Gradient Reversal Layer (GRL) and used a common encoder to align the source and target feature spaces. 
Alternately, Tzeng et al. \cite{tzeng2017adversarial} first trained an deep network on the labelled source data, then used an adversarial domain discriminator to train a separate target data encoder, thus decoupling the encoders for the source and target domains. 
Such global domain-wise adversarial alignment, however, runs the risk of losing the intrinsic class-wise discriminability in the embedding space for the target domain. 
Li et al. \cite{li2019joint} proposed to simultaneously solve one domain-specific and one class-specific mini-max problems, in an effort to preserve class-wise discriminability during adversarial alignment.
Structural regularization between the source and target class clusters in the feature space \cite{pan2019transferrable, tang2020unsupervised} has also been explored as a means for non-adversarial domain alignment. 

Domain divergence minimization, adversarial latent space alignment, cluster structure regularization, etc. strategies require both source and target data to be accessible during the adaptation process, giving rise to concerns about privacy and security for the source data, as well as extra data storage requirements for edge applications. 
Source-free domain adaptation \cite{chidlovskii2016domain, shot, yang2021generalized, ding2022source} was proposed to address such cases, where a source model is first trained on the source data, and then instead of the source data itself, the source encoder and classifiers are transferred for subsequent target domain adaptation. 
Chidlovskii et al. \cite{chidlovskii2016domain} proposed a semi-SFDA method where instead of using the entire source domain data, only prototypes or few examples for each source class was used for adaptation. 
Liang \textit{et al.} \cite{shot} proposed to lock the source hypothesis or classifier, and adapt the source encoder to the target data by means of self-training using class centroid-based pseudolabel refinement and information maximization \cite{krause2010discriminative, shi2012information}. 
Yang et al. \cite{yang2021generalized} argued that neighboring target samples should also be consistent in their pseudolabel assignment, and applied consistency regularization among neighboring samples as an auxiliary pseudolabel refinement technique. 
Ding et al. \cite{ding2022source} proposed a source distribution estimation technique, and sampled from the estimated source distribution to contrastively align with the target data distribution. 

\subsection{Continual Domain Adaptation}

Continual learning is the process of learning new tasks with the same neural network over different episodes or training stages. 
While mammals can expand their learning capacity to fit new tasks, artificial neural networks, due to their rigid connectionist structures, often fail to remember knowledge learned from earlier tasks, and succumb to catastrophic forgetting \cite{mccloskey1989catastrophic, robins1995catastrophic} of the earlier tasks when trained on a new task. 
Inspired by synaptic consolidation in neurons of humans and other animals, Kirkpatrick et al. \cite{kirkpatrick2017overcoming} and Zenke et al. \cite{zenke2017continual} proposed to regularize the network parameter updates to preserve the weights important for previously learned tasks. 
Similarly, inspired by the role of sleep in memory consolidation \cite{ji2007coordinated, walker2004sleep}, several methods \cite{rebuffi2017icarl, hayes2020remind} propose selective replay of old samples from earlier tasks, along with new data during training on a new task to overcome catastrophic forgetting. 
Taking cues from memory replay, Taufique et al. \cite{taufique2023condaTAI} proposed to conduct continual domain adaptation on static target domains by selectively storing raw samples from earlier target data batches, and replaying them with new samples from the current target data batch, along with self-training with information maximization and self-supervised pseudolabeling. 
In a similar approach, Taufique et al. \cite{taufique2022unsupervised} proposed carry out continual DA on gradually varying or dynamic target domains by minimizing an additional contrastive loss. 

In this paper, we evaluate two memory replay based continual DA models ConDA \cite{taufique2023condaTAI} and UCL-GV \cite{taufique2022unsupervised} on four benchmark gradually degrading weather datasets that we generate. As a baseline, we run the standard DA model SHOT \cite{shot} in a continual manner, where a model is fed the target data in small batches, and the model cannot get access to a data batch that has earlier been adapted to or learned from. A brief description of the models is given in Section \ref{sec:method}.

\section{Method}
\label{sec:method}
We utilize the standard source-free DA model SHOT \cite{pmlr-v119-liang20a} under the continual DA framework, and call this model Continual-SHOT henceforth in this paper. 
We further evaluate two state-of-the-art continual DA models ConDA \cite{taufique2023condaTAI} and UCL-GV \cite{taufique2022unsupervised}. 
For all our models, we utilize the standard RedNet50 \cite{He_2016_CVPR} backbone and additionally consider backbones based on vision transformers \cite{vit} and Shifted Window (SWIN) transformers \cite{swin} to evaluate their performance.

\begin{figure}
\centering
\begin{minipage}[b]{1\linewidth}
  \centering
  \centerline{\includegraphics[width=\textwidth]{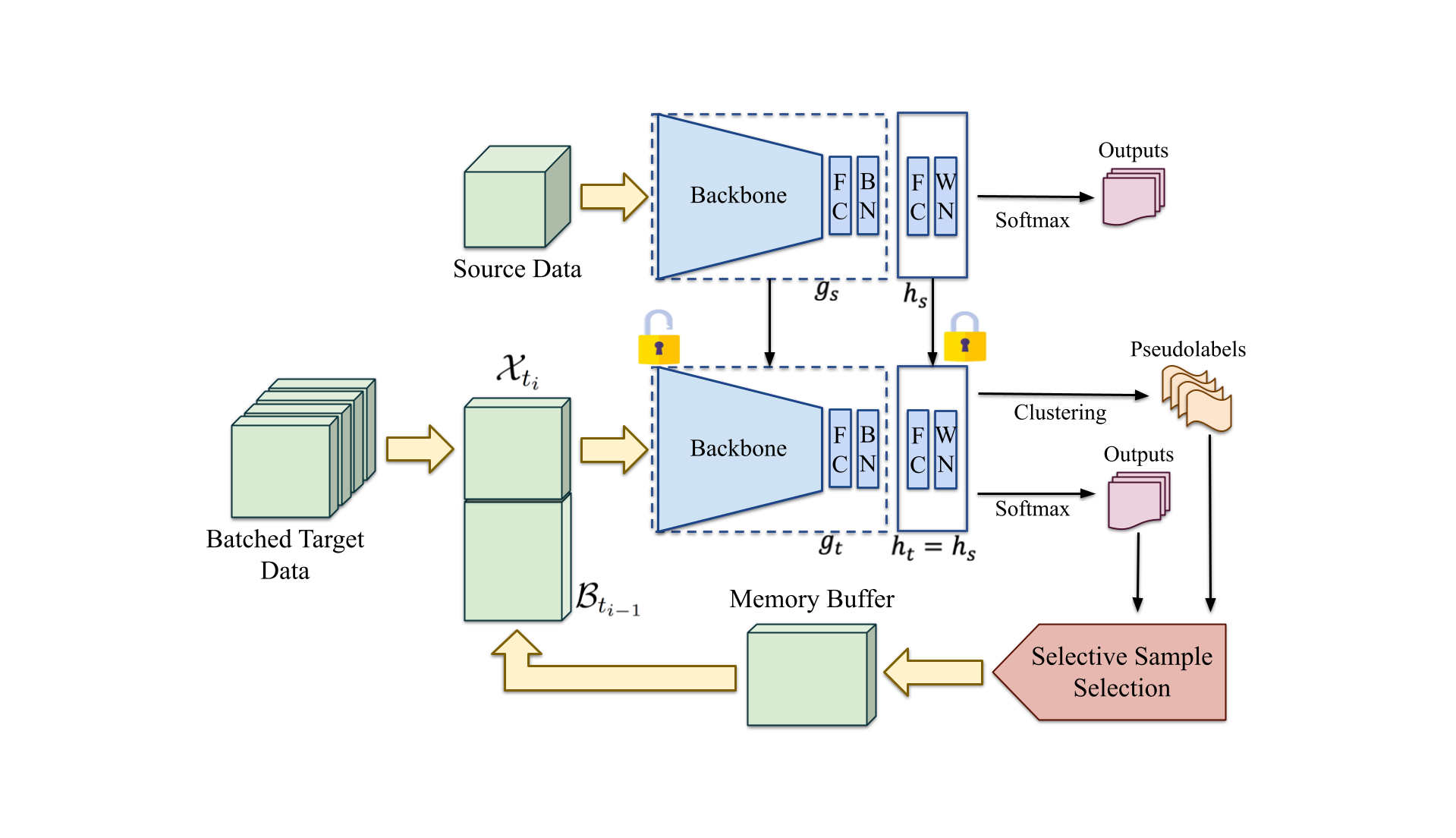}}
  \label{fig:conda_ucl}
 \end{minipage}
\caption{Overview of operation of source-free and buffer-fed continual DA models ConDA and UCL-GV. The source encoder and classifier are transferred, but only the encoder is unlocked for adaptation. The target samples are fed  to the model in small batches, and after adaptation on that batch, only certain samples are selectively stored in a memory buffer and are replayed with the next incoming batch of target data.}
\label{fig:conda_ucl}
\end{figure}

An illustration of the general operation of continual models ConDA and UCL-GV is given in Figure \ref{fig:conda_ucl}. 
Both ConDA and UCL-GV use a memory buffer to selectively store target samples from previously encountered batches, and then replay them with new incoming target data. Continual-SHOT does not have a buffer and does not utilize any memory replay. 

The source domain with $\mathcal{C}_s$ classes is denoted by $\mathcal{D}_s$, where the source data $\mathcal{X}_s$ with $n_s$ labeled samples is represented by $\{x_s^z, y_s^z\}_{z=1}^{n_s}$, where $x_s \in \mathcal{X}_s$ and labels $y_s \in \mathcal{Y}_s$. 
Similar to the formulation of continually varying domains in Kumar et. al \cite{kumar2020understanding} and Taufique et al. \cite{taufique2022unsupervised}, an unlabeled intermediate domain $\mathcal{D}_{int}$, with $\mathcal{C}_{int}$ classes and $\mathcal{X}_{int}$ samples is considered in addition to an unlabeled target domain, $\mathcal{D}_{tar}$, that has $\mathcal{C}_{tar}$ classes and samples $\mathcal{X}_{tar}$. 
The intermediate domain $\mathcal{D}_{int}$ and target domain $\mathcal{D}_{tar}$ are combined to form the domain $\mathcal{D}_t$ with the unlabeled data $\mathcal{X}_t = \mathcal{X}_{int} \bigcup \mathcal{X}_{tar}$, such that $\{x_t^z\}_{z=1}^{n_t} \in \mathcal{X}_t$ where $n_t$ is the total number of unlabeled samples.
Under our use case of closed set adaptation, we consider the classes to be the same across all domains. 
Therefore, $\mathcal{C}_t = \mathcal{C}_{int}=\mathcal{C}_{tar} = \mathcal{C}_s$. 
Domain $\mathcal{D}_t$ is split into $m$ sequential batches of a fixed sample size, such that $\mathcal{X}_t = \{\mathcal{X}_{t_1}, \mathcal{X}_{t_2}, \mathcal{X}_{t_3}, \cdot \cdot \cdot, \mathcal{X}_{t_m}\}$ and $n_t = \sum_{i=1}^m n_{t_i}$, where 
$i = 1, 2, 3, \cdot \cdot \cdot, m$ are the time steps at which the batches of gradually degrading data are encountered. 
The purpose of source-free continual domain adaptation on gradually varying domains is to train a source model $f_{s}(\theta_s): \mathcal{X}_s \rightarrow \mathcal{Y}_s$, and adapt it to $\mathcal{D}_t$ continually, so that the model $f_{t}(\theta_t): \mathcal{X}_{t_i} \rightarrow \mathcal{Y}_{t_i}$ provides better performance on target data $\mathcal{X}_{tar}$ at the end of adaptation when $i=m$, compared to that of $f_{s}(\theta_s)$ on $\mathcal{X}_{tar}$.

The architectures of Continual-SHOT, ConDA and UCL-GV generally follow a structure that is similar to SHOT \cite{shot} with the addition of a buffer for memory replay in ConDA and UCL-GV. 
Each model has two parts: 
$(i)$ a feature extractor part $g$ and  $(ii)$ a classifier or hypothesis part $h$. The feature extractor includes the backbone, followed by a bottleneck layer which consists of a fully connected layer with 256 output nodes and a batch normalization layer. 
The feature extractor backbone is typically ResNet50 \cite{He_2016_CVPR}, but in this paper we also explore ViT \cite{vit} and Swin \cite{swin} transformer architectures.  
ViT breaks an image into patches of $16\times16$ pixels, embeds each patch into a token and then applies self-attention across each of the token pairs. 
This enables ViT to achieve global contextual understandings of an image, even at the shallower blocks, whereas CNN-based architectures are hierarchical and achieve global feature representations in the deeper layers.
On the other hand, SWIN breaks the image into a number of windows, and each window is then divided into a number of smaller $4\times4$ pixel patches. Swin applies self-attention more locally, across only the patches within each window. Swin overcomes local feature representations by shifting the windows, and achieves hierarchical feature scale by patch merging. 
The classifier or hypothesis part of our model consists of a fully connected layer and a weight normalization layer. 
The source model undergoes supervised training with label-smoothing for softer class boundaries.

The source trained model is transferred to the target model where the classifier $h_t = h_s$ is locked, and the target feature extractor $g_t$ is initialized with the source feather extractor $g_s$. 
During target adaptation, ConDA and UCL-GV utilizes a memory buffer $\mathcal{B_t}$ of length $\mathcal{L}$. 
The buffer maintains a class-balance across all classes, i.e the buffer is split into $C_t$ sections of equal size $\mathcal{L}/C_t$. 
Let at time step $i$ 
, the current buffer state be represented by $B_{t_{i-1}}$, consisting of selectively stored samples from time step $i-1$. 
At time step $i$, the incoming target data $\mathcal{X}_{t_i}$ is combined with $B_{t_{i-1}}$, and the model is adapted on $\mathcal{X}_{t_i}^\prime = \mathcal{X}_{t_i} \bigcup B_{t_{i-1}}$. 
Pseudolabels $y_{t_i}^\prime \in \mathcal{Y}_{t_i}^\prime$ are calculated for $x_{t_i}^\prime \in \mathcal{X}_{t_i}^\prime$ from the model predictions, and then refined using a self-supervised clustering algorithm \cite{shot}. 

All the continual models, Continual-SHOT, ConDA, and UCL-GV, are adapted with a cross-entropy loss against the refined pseudolabels $y_{t_i}^\prime$, and variants of the information maximization (IM) loss \cite{shot, taufique2023condaTAI,taufique2022unsupervised}.
To account for the class-balanced structure of the buffer, the diversity loss of the IM loss in SHOT \cite{shot} is modified to equal diversity loss \cite{taufique2023condaTAI, taufique2022unsupervised} in ConDA and UCL-GV.
UCL-GV is adapted with an additional prototypical contrastive loss \cite{li2020prototypical} between the class prototypes or cluster centers for buffer samples $B_{t_{i-1}}$, and the incoming samples $\mathcal{X}_{t_i}$. 

At the end of adaptation for time step $i$, the buffer is repopulated with $\mathcal{X}_{t_i}$. 
To maintain class-balance in the buffer, if any class has more than $\mathcal{L} / C_t$, ConDA selects samples based on confidence score, while UCL-GV selects samples randomly to store in the memory buffer.   
The remaining buffer slots are filled with $B_{t_{i-1}}$ following the same strategy, to form the buffer state $B_{t_{i}}$ for use in time step $i+1$. 
Adaptation continues until all the incoming target batches are processed, and the adaptation accuracy of the final model on the target data $\mathcal{X}_{tar}$ is taken as the result.

Gradually varying domains include the initial source domain, intermediate domains obtained during progressively varying transitions, and the final target domain. 
During adaptation to gradually varying domains, the source-trained model adapts first to the least degraded intermediate domain, and then to increasingly degraded intermediate domains, and finally to the highest degraded target domain. The proposed datasets for gradually varying domain adaptation are discussed in the next section.

\section{Benchmark Datasets}
\label{sec:dataset}
To the best of our knowledge, no existing dataset meets our criteria for evaluating continual domain adaptation under gradually changing weather conditions. We therefore utilize two existing aerial datasets AID \cite{xia2017aid} and UCM \cite{yang2010bag} to generate gradually varying weather conditions using the 
{\em imgaug} 
Python library \cite{imgaug}. We use all 30 classes for AID, and all 21 classes for UCM. We use 2 augmenters {\em CloudLayer} and {\em SnowflakesLayer} from {\em imgaug.augmenters.weather} library to synthesize cloudy, and snowfall weather conditions on real AID, and UCM images. With 2 augmentations on AID, and 2 augmentations on UCM, we get a total of 4 datasets with gradually degrading weather conditions. We call the new AID dataset with cloud cover distortion, and with snowfall distortion AID-CC and AID-SF, respectively. Similarly, we name the new UCM dataset with cloud cover distortion, and with snowfall distortion UCM-CC and UCM-SF, respectively.

\begin{figure}
\centering
\begin{minipage}[th]{0.95\linewidth}
  \centering
  \centerline{\includegraphics[width=\textwidth]{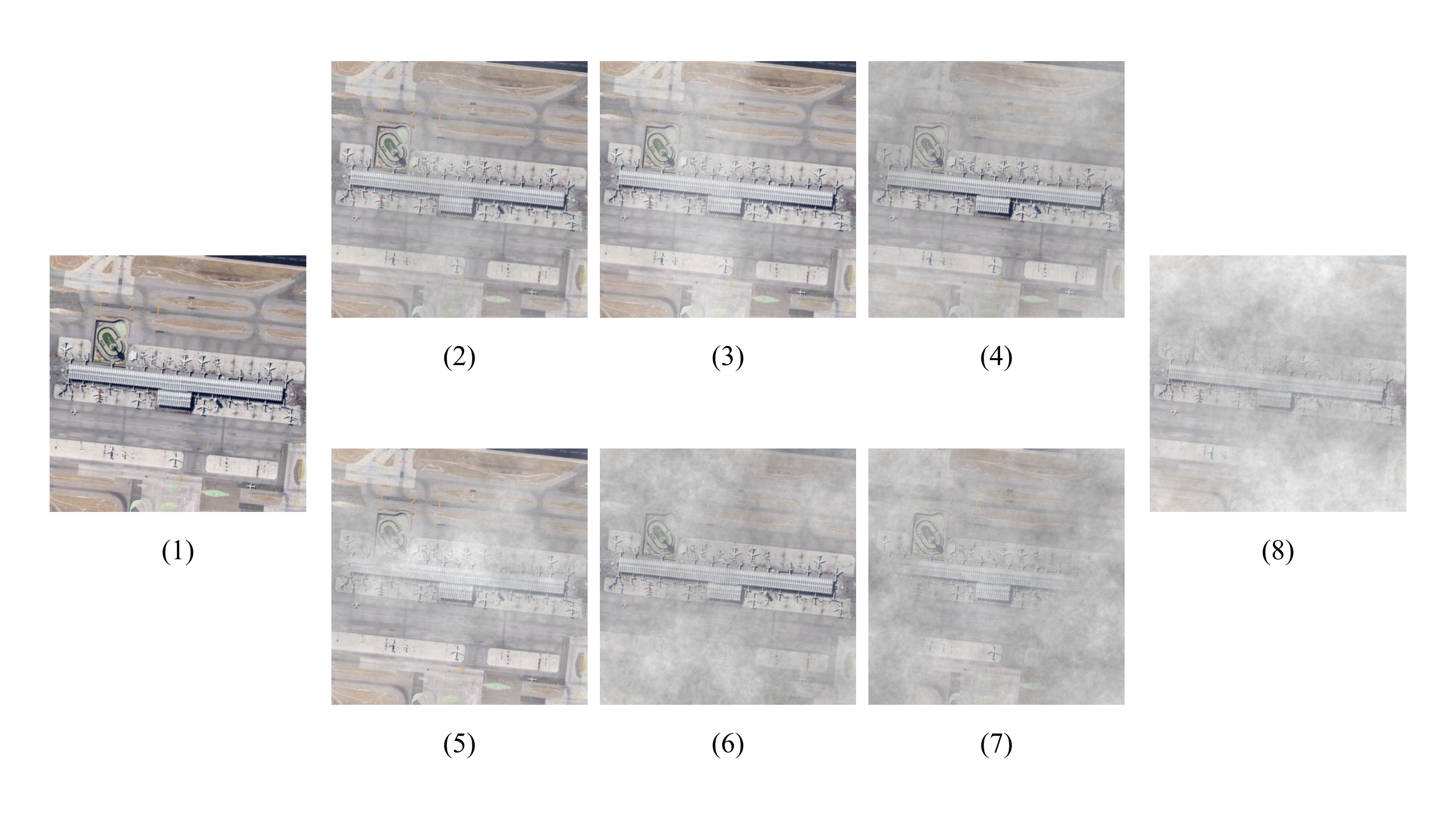}}
  \label{fig:aid_cloud}
 \end{minipage}
\caption{AID-CC dataset with cloud cover degradation. (1) is the source domain, (8) is the target domain, and (2-7) are progressively degrading intermediate domains.}
\label{fig:aid_cloud}
\end{figure}

\begin{figure}
\centering
\begin{minipage}[bh]{0.95\linewidth}
  \centering
  \centerline{\includegraphics[width=\textwidth]{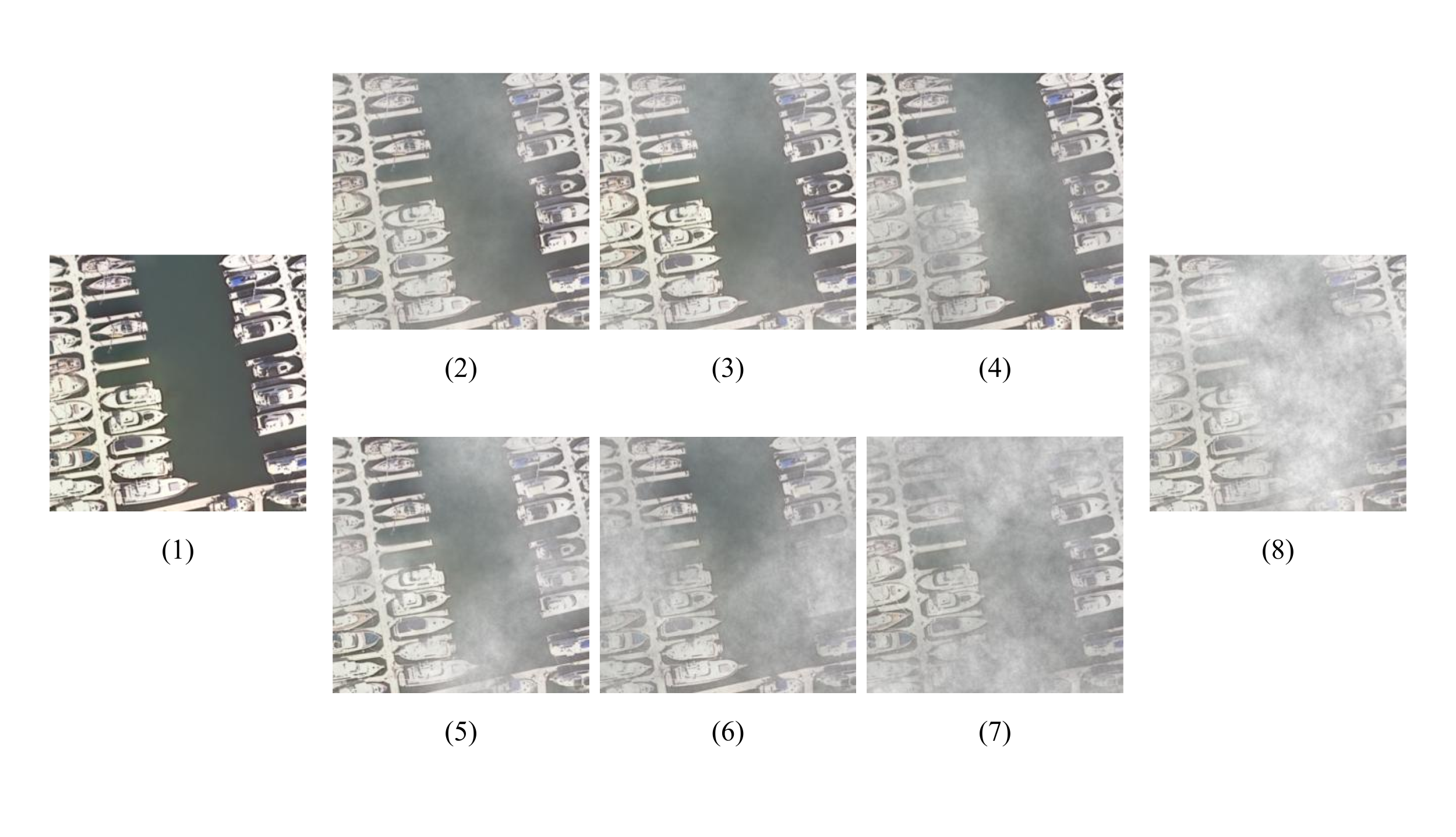}}
  \label{fig:ucm_cloud}
 \end{minipage}
\caption{UCM-CC dataset with cloud cover degradation. (1) is the source domain, (8) is the target domain, and (2-7) are progressively degrading intermediate domains.}
\label{fig:ucm_cloud}
\end{figure}

\begin{figure}
\centering
\begin{minipage}[th]{0.95\linewidth}
  \centering
  \centerline{\includegraphics[width=\textwidth]{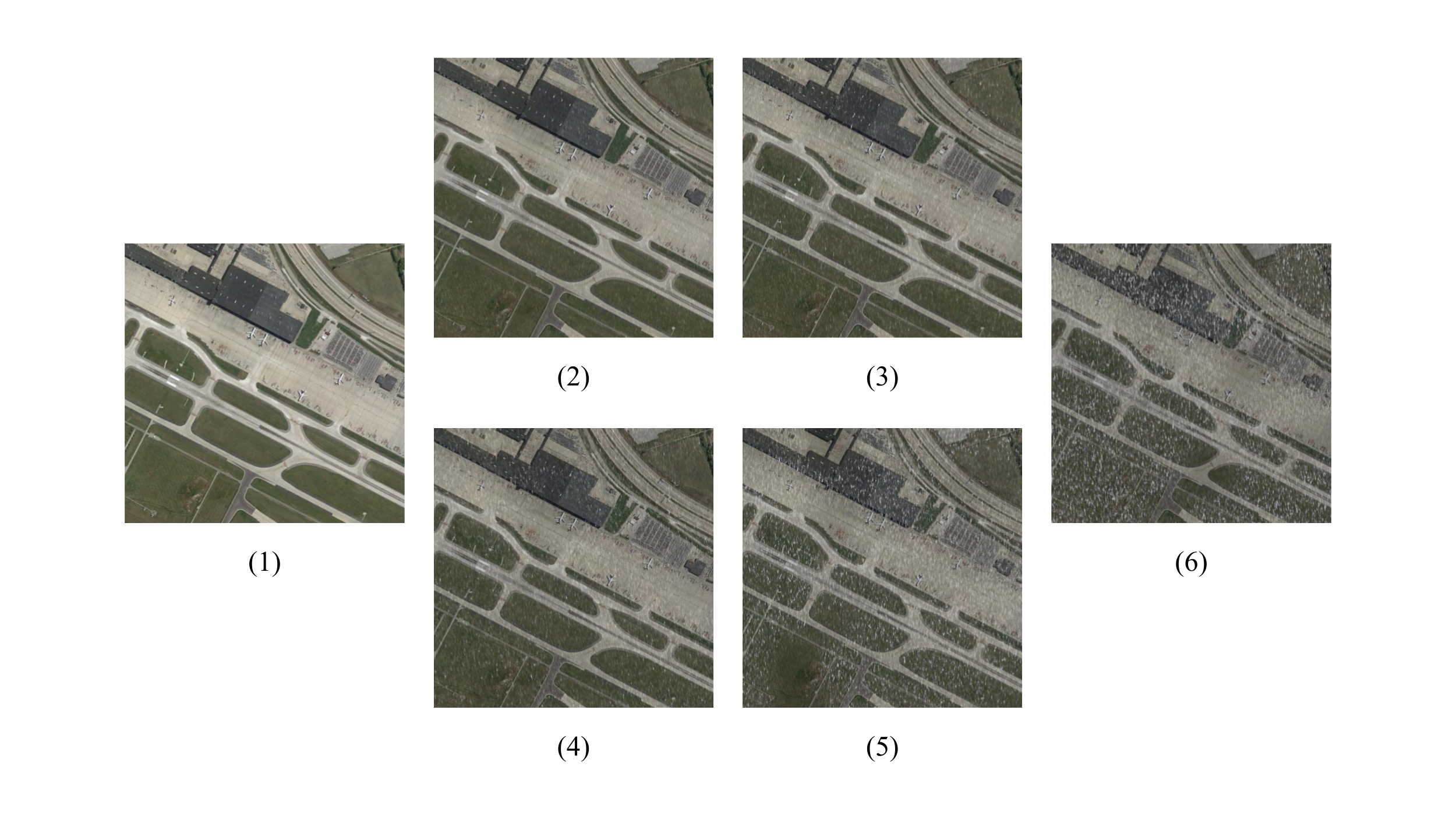}}
  \label{fig:aid_snowflake}
 \end{minipage}
\caption{AID-SF dataset with snowfall degradation. (1) is the source domain, (6) is the target domain, and (2-5) are progressively degrading intermediate domains.}
\label{fig:aid_snowflake}
\end{figure}

\begin{figure}
\centering
\begin{minipage}[bh]{0.95\linewidth}
  \centering
  \centerline{\includegraphics[width=\textwidth]{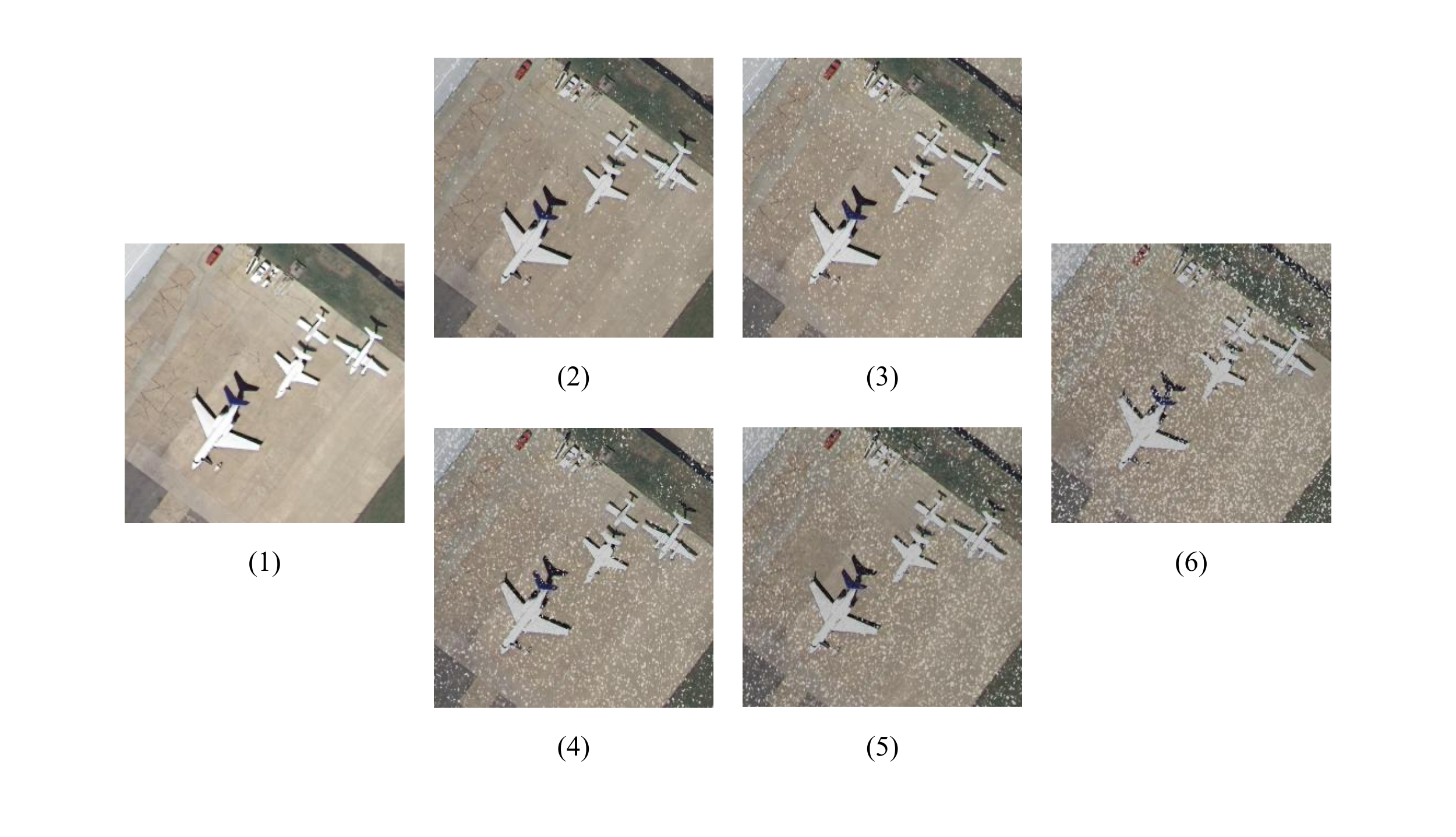}}
  \label{fig:ucm_snowflake}
 \end{minipage}
\caption{UCM-SF dataset with snowfall degradation. (1) is the source domain, (6) is the target domain, and (2-5) are progressively degrading intermediate domains.}
\label{fig:ucm_snowflake}
\end{figure}

We take the clear weather images from AID, and UCM as the source data for each respective dataset. 
The 7 levels of cloud cover degradation are made by varying the density and size of clouds. 
The 5 levels of snowfall degradation are made by varying the density of the snowflakes, and overall brightness of the scene. 
The data with the highest level of degradation for both types are taken as the respective target domain data, while the rest are treated as gradually varying intermediate domains, depending on the intensity of weather degradation. 
We therefore have 6 intermediate domains for cloud cover, and 4 intermediate domains for snowfall. 

We present a few examples of our newly created   synthetic datasets in Figures \ref{fig:aid_cloud}, \ref{fig:ucm_cloud}, \ref{fig:aid_snowflake}, and \ref{fig:ucm_snowflake}. Figures \ref{fig:aid_cloud} and \ref{fig:ucm_cloud} show examples of the 7 stages of gradually worsening cloud coverage on AID and UCM, respectively. 
Figures \ref{fig:aid_snowflake} and \ref{fig:ucm_snowflake} show examples of the 5 gradually degrading snowfall conditions on AID and UCM, respectively. 

\section{Implementation Details}

All three models we evaluated consist of a backbone or feature extractor, followed by a bottleneck layer, and finally a classifier layer. 
The buffer sizes for ConDA and UCL-GV are fixed at 420 samples, while Continual SHOT does not contain a buffer.
The models are adapted with an SGD optimizer with momentum of $0.9$. 
The initial learning rates $\eta_0 = 0.02$ and $\eta_0 = 0.002$ are used with a learning rate scheduler such that the learning rate $\eta = \eta_0 \cdot (1 + 10 \cdot p)^{-0.75}$, where $p=\frac{i}{i_T}$ changes from 0 to 1 for each incoming batch. Here $i$ is the current iteration, and $i_T$ is the total number of iterations for each incoming batch of intermediate/target data.

\section{Results and Discussion}

\begin{table}[th]
\caption{Initial results on the gradually degrading AID and UCM datasets with \textit{ResNet-50} backbone on the final target domain. Source-trained refers to the model trained on the source data only, without any adaptation. The top accuracy is in bold and the second best is underlined.}\label{tab:results_res}
\begin{center}
\begin{tabular}{|c|P{1.8cm}|P{1.8cm}|P{1.8cm}|P{1.8cm}|}
    \hline
    \multirow{2}{*}{Method} & 
    \multicolumn{2}{c|}{Cloud Cover} & 
    \multicolumn{2}{c|}{Snowfall} 
    \\ 
    \cline{2-5}
    & AID-CC & UCM-CC & AID-SF & UCM-SF  \\ 
    \hline
    Source-trained  & 12.29 & 32.81 & 42.65 & 58.38 \\
    Continual-SHOT \cite{pmlr-v119-liang20a}  & \textbf{84.14} & 80.10 & 94.21 & 95.00 \\
    ConDA \cite{taufique2023condaTAI}  & \underline{80.41} & \textbf{85.54} & \underline{95.38} & \underline{95.90}  \\
    UCL-GV \cite{taufique2022unsupervised}  & 79.40 & \underline{85.19} & \textbf{95.49} & \textbf{95.95}  \\
    \hline
\end{tabular}    
\end{center}
\end{table}

Initial results for all three methods considered 
using a ResNet-50 backbone are shown in Table \ref{tab:results_res}.
The two continual DA models, UCL-GV and ConDA, outperform Continual-SHOT for snowfall degradation, but the results for cloud cover are mixed.  
For AID-CC, we see significant drops in performance for the ConDA and UCL-GV, compared to continual-SHOT. 
With additional examination of the results, 
we found occasional lack of stability that ConDA and UCL-GV may encounter for certain batches during adaptation.
The continual batches within a domain do not have any particular order in which they are received, and the performance drop can occur at any time during the adaptation process. 
Such adaptation instability needs to be addressed to improve performance, since continual DA does not revisit batches of images that have already been processed or seen by the model.

\begin{figure*}
  \centering
  \begin{subfigure}{0.9\linewidth}
    \includegraphics[width=1\textwidth]{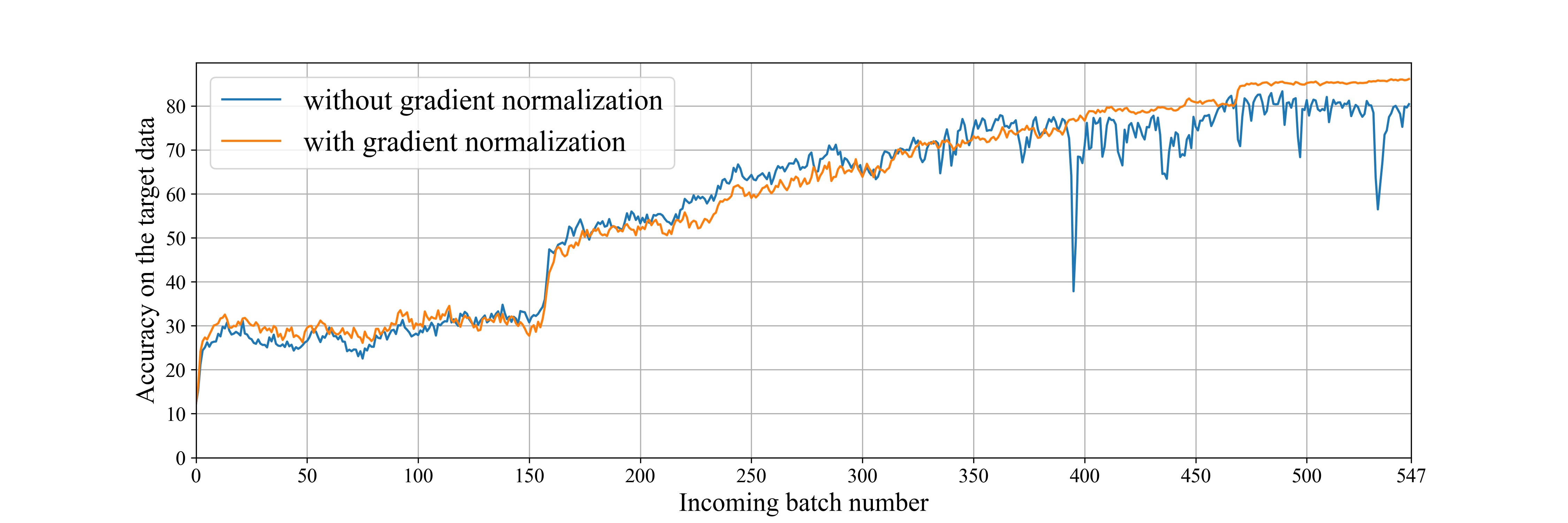}
    \caption{Accuracy of ConDA with \textit{ResNet-50} backbone on the final target data, with and without gradient normalization, as it continually adapts to the incoming batches of intermediate and target domains of AID-CC.}
    \label{fig:fluctuations-conda}
  \end{subfigure}
  \hfill
  \begin{subfigure}{0.9\linewidth}
    \includegraphics[width=1\textwidth]{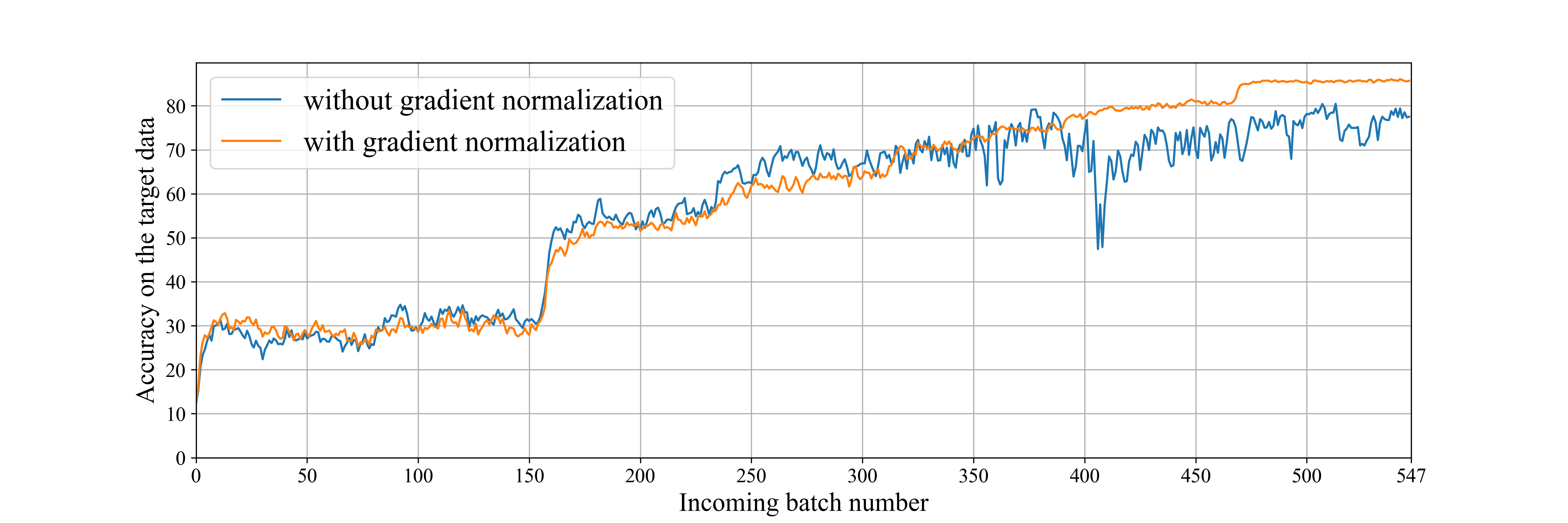}
    \caption{Accuracy of UCL-GV with \textit{ResNet-50} backbone on the final target data, with and without gradient normalization, as it continually adapts to the incoming batches of intermediate and target domains of AID-CC.}
    \label{fig:fluctuations-ucl}
  \end{subfigure}
  \caption{Effect of gradient normalization on adaptation stability for the continual models ConDA and UCL-GV.}
  \label{fig:fluctuations}
\end{figure*}

We propose gradient normalization to help stabilize the adaptation process for the continual models, and improve their performance.
Empirically, we conduct $L_2$-normalization of all the gradients after backpropagation through the model and before optimization for each adaptation iteration.
In Figure \ref{fig:fluctuations}, we plot the continual adaptation performance for each incoming batch, with and without gradient normalization, for the continual models on AID-CC. 


\begin{table}[th]
\caption{Results on the gradually degrading AID and UCM datasets, using ResNet-50 backbone, gradient normalization, and initial learning rate of $\eta_0 = 0.002$ and $\eta_0 = 0.02$.
Source-trained refers to the model trained on the source data only, without any continual target adaptation over the intermediate and target domains. The top accuracy is in bold and the second best is underlined.} 
\label{tab:results_res_normalized}
\begin{center}
\begin{tabular}{|c|P{1.1cm}|P{1.1cm}|P{1.1cm}|P{1.1cm}|P{1.1cm}|P{1.1cm}|P{1.1cm}|P{1.1cm}|}
    \hline
    \multirow{2}{*}{ Method} & 
    \multicolumn{2}{c|}{AID-CC} &
    \multicolumn{2}{c|}{UCM-CC} &
    \multicolumn{2}{c|}{AID-SF} &
    \multicolumn{2}{c|}{UCM-SF}
    \\
    \cline{2-9}
    & $\eta_0 = 0.002$ & $\eta_0 = 0.02$ & $\eta_0 = 0.002$ & $\eta_0 = 0.02$ & $\eta_0 = 0.002$ & $\eta_0 = 0.02$ & $\eta_0 = 0.002$ & $\eta_0 = 0.02$ \\ 
    \hline
    Source-trained &  12.29 & 12.29 & 32.81 & 32.81 & 42.65 & 42.65 & 58.38 & 58.38 \\
    Continual-SHOT \cite{pmlr-v119-liang20a} & 80.37 & 79.36 & 67.95 & 81.71 & 88.71 & 94.50 & 78.29 & 95.57  \\
    ConDA \cite{taufique2023condaTAI} & \textbf{86.13} & 58.91 & 79.43 & \underline{81.33} & \underline{94.87} & 93.12 & 90.05 & \textbf{97.38} \\
    UCL-GV \cite{taufique2022unsupervised} & \underline{85.67} & 65.09 & 78.76 & \textbf{82.10} & \textbf{95.05} & 93.68 & 90.48 & \underline{96.71} \\
    \hline
\end{tabular}
\end{center}
\end{table}

From Fig. \ref{fig:fluctuations}, we can see that ConDA and UCL-GV may face significant stability issues at times, and model performance may  drop significantly (up to $\sim 40\%$) from one incoming batch to the next.
Although the continual models start to recover in the subsequent incoming batches for AID-CC, they may not adapt optimally.
In our experiments, it is evident that gradient normalization greatly mitigates the drops in performance for some of the batches, that are observed in the original forms of the models. With gradient normalization, both ConDA and UCL-GV gradually continue to better adapt to the target domain as the adaptation process progresses.


The results for all 3 methods with gradient normalization, and different learning rates are tabulated in Table \ref{tab:results_res_normalized}. 
In all cases, ConDA and UCL-GV outperform continual SHOT, but the best results depend on the learning rate.
To understand the effects of the learning rate, it is noted that for optimal performance, continual DA necessitates the models to undergo fast optimization, because revisits to earlier data batches are not allowed, and the models need to adapt to the target domain over single passes of the continual data stream.
On the smaller UCM-CC and UCM-SF datasets, both ConDA and UCL-GV adapt better with a higher learning rate, due to faster optimization afforded by the higher initial learning rate.  
On the other hand, the models cannot adequately adapt to the target domain for UCM-CC and UCM-SF at the smaller initial learning rate of $\eta=0.002$, due to comparatively slower optimization. 
Therefore, for UCM-CC and UCM-SF, best results are obtained for higher learning rates.

However, the continual adaptation process becomes more susceptible to instability at higher learning rates, the effects of which can be seen in the results for AID-CC at $\eta_0 = 0.02$.
On the larger AID-CC and AID-SF datasets, better performance is obtained with the smaller initial learning rate of $\eta_0=0.002$. 
As AID-CC and AID-SF have large number of samples to process, continual models can reach an optimal solution with a slower optimization at smaller learning rates, and avoid the risk of higher instability at higher learning rates.

We further evaluate the continual models with two transformer backbones: Vision Transformer (ViT) \cite{vit} and Swin-V2 \cite{swin, swinv2}. 
To keep the computational load tractable, we choose the base versions of the transformer backbones for our experiments. 
We report the adaptation performance of the three models, with and without gradient normalization, on 
the AID-CC and AID-SF datasets at the lower initial learning rate of $\eta_0 = 0.002$
in Table \ref{tab:res_transformers1}, and 
on the UCM-CC and UCM-SF datasets at the higher initial learning rate of $\eta_0 = 0.002$
in Table \ref{tab:res_transformers2}.

\begin{table}[t!]
\caption{Results on AID-CC, and AID-SF with ResNet-50, ViT-B, and Swin-B backbones, at initial learning rate of 0.002. Source-trained method refers to the model trained on the source data only, without any continual target adaptation over the intermediate and target domains. The top accuracy is in bold and the second best is underlined. respectively.}\label{tab:res_transformers1}
\begin{center}
\begin{tabular}{|c|c|*{4}{P{2.0cm}|}}
    \hline
    \multirow{2}{*}{Method} & 
    \multirow{2}{*}{\makecell{\parbox{1.9cm}{Backbone (\# params)}}} &
    \multicolumn{2}{c|}{AID-CC} & 
    \multicolumn{2}{c|}{AID-SF} \\
    \cline{3-6}
    & & w/o. Grad Norm & w. Grad Norm & w/o. Grad Norm & w. Grad Norm\\ 
    \hline
    Source-trained & \multirow{4}{*}{\parbox{1.9cm}{ResNet-50 (23M)}} & 12.29 & 12.29 & 42.65 & 42.65\\
    Continual-SHOT \cite{pmlr-v119-liang20a} &  & 84.14 & 80.37 & 94.21 & 88.71 \\
    ConDA \cite{taufique2023condaTAI}  &  & 80.41 & 86.13 & 95.38 & 94.87 \\
    UCL-GV \cite{taufique2022unsupervised} &  & 79.40 & 85.67 & 95.49 & 95.05 \\
    \hline
    Source-trained & \multirow{4}{*}{\parbox{1.9cm}{ViT-B (86M)}} & 11.50 & 11.50 & 55.85 & 55.85 \\ 
    Continual-SHOT \cite{pmlr-v119-liang20a} &  & 80.43 & 74.18 & 90.41 & 88.79 \\
    ConDA \cite{taufique2023condaTAI}  &  & 78.28 & 79.94 & 89.65 & 89.77 \\
    UCL-GV \cite{taufique2022unsupervised} &  & 79.55 & 79.84 & 88.56 & 90.29 \\
    \hline
    Source-trained & \multirow{4}{*}{\parbox{1.9cm}{Swin-B (88M)}} & 19.96 & 19.96 & 67.34 & 67.34 \\
    Continual-SHOT \cite{pmlr-v119-liang20a} &  & 89.76 & 91.20 & 96.29 & 94.12 \\
    ConDA \cite{taufique2023condaTAI}  &  & 81.82 & \textbf{93.20} & 95.82 & \underline{97.77} \\
    UCL-GV \cite{taufique2022unsupervised} &  & 81.67 & \underline{92.82} & 93.22 & \textbf{97.84} \\
    \hline    
\end{tabular}
\end{center}
\end{table}

\begin{table}[th]
\caption{Results on UCM-CC, and UCM-SF with ResNet-50, ViT-B, and Swin-B backbones, at initial learning rate of 0.02. Source-trained method refers to the model trained on the source data only, without any continual target adaptation over the intermediate and target domains. The top accuracy is in bold and the second best is underlined. respectively.}\label{tab:res_transformers2}
\begin{center}
\begin{tabular}{|c|c|*{4}{P{2.0cm}|}}
    \hline
    \multirow{2}{*}{Method} & 
    \multirow{2}{*}{\makecell{\parbox{1.9cm}{Backbone (\# params)}}} &
    \multicolumn{2}{c|}{UCM-CC} & 
    \multicolumn{2}{c|}{UCM-SF} \\
    \cline{3-6}
    & & w/o. Grad Norm & w. Grad Norm & w/o. Grad Norm & w. Grad Norm\\ 
    \hline
    Source-trained & \multirow{4}{*}{\parbox{1.9cm}{ResNet-50 (23M)}} & 32.81 & 32.81 & 58.38 & 58.38 \\
    Continual-SHOT \cite{pmlr-v119-liang20a} &  & 9.33 & 81.71 & 73.81 & 95.57 \\
    ConDA \cite{taufique2023condaTAI}  &  & 6.95 & 81.33 & 96.00 & \textbf{97.38} \\
    UCL-GV \cite{taufique2022unsupervised} &  & 8.52 & 82.10 & 92.14 & 96.71 \\
    \hline
    Source-trained & \multirow{4}{*}{\parbox{1.9cm}{ViT-B (86M)}} & 48.00 & 48.00 & 59.24 & 59.24 \\ 
    Continual-SHOT \cite{pmlr-v119-liang20a} &  & 82.00 & 79.38 & 95.10 & 94.14 \\
    ConDA \cite{taufique2023condaTAI}  &  & 56.29 & 77.81 & 82.71 & 95.71 \\
    UCL-GV \cite{taufique2022unsupervised} &  & 25.52 & 71.95 & 76.19 & 94.52 \\
    \hline
    Source-trained & \multirow{4}{*}{\parbox{1.9cm}{Swin-B (88M)}} & 14.10 & 14.10 & 72.67 & 72.67 \\
    Continual-SHOT \cite{pmlr-v119-liang20a} &  & 55.62 & \underline{86.05} & 87.10 & \underline{97.05} \\
    ConDA \cite{taufique2023condaTAI}  &  & 3.76 & \textbf{87.29} & 69.43 & 96.95 \\
    UCL-GV \cite{taufique2022unsupervised} &  & 7.38 & 85.81 & 46.38 & \textbf{97.38} \\
    \hline    
\end{tabular}
\end{center}
\end{table}

\begin{figure*}
  \centering
  \begin{subfigure}{0.9\linewidth}
    \includegraphics[width=1\textwidth]{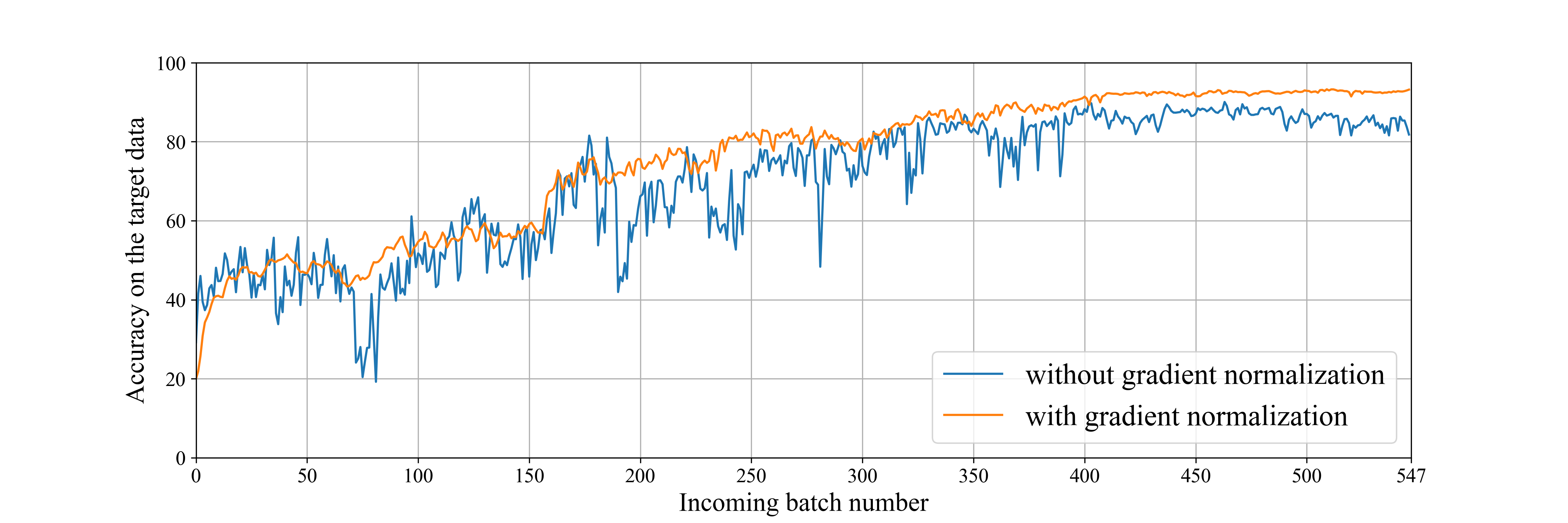}
    \caption{Accuracy of ConDA with \textit{Swin} backbone, both with, and without gradient normalization on the final target data as it continually adapts to the incoming batches of intermediate and target domains of AID-CC.}
    \label{fig:fluctuations-conda}
  \end{subfigure}
  \hfill
  \begin{subfigure}{0.9\linewidth}
    \includegraphics[width=1\textwidth]{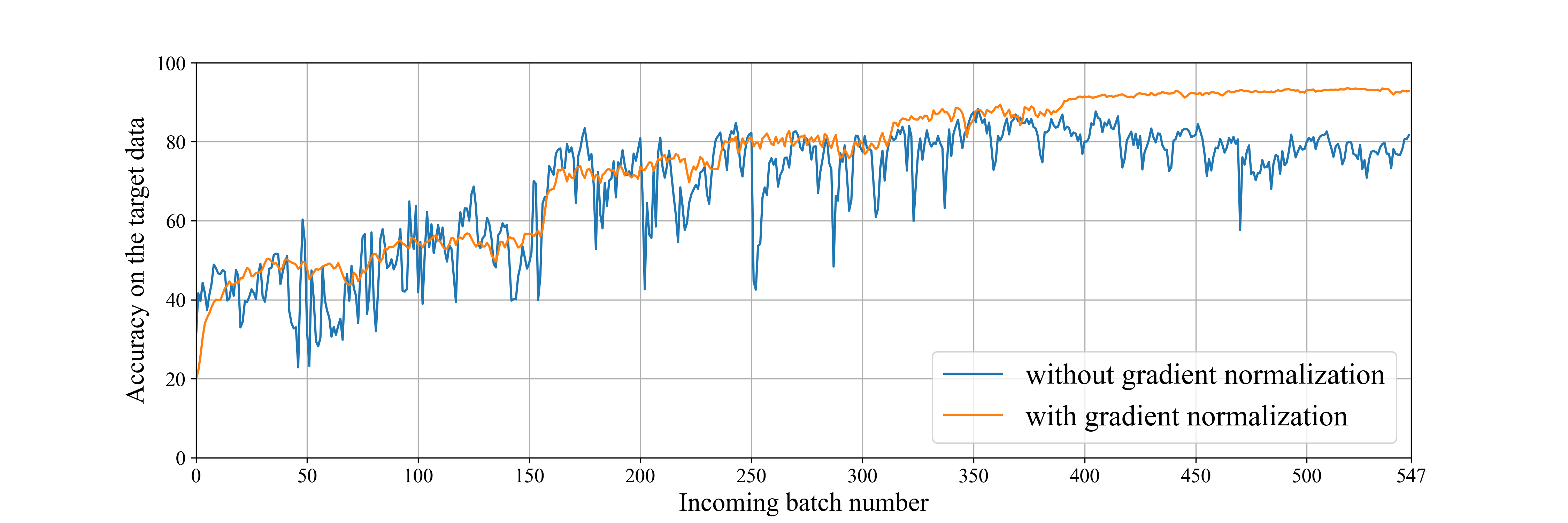}
    \caption{Accuracy of UCL-GV with Swin backbone, both with, and without gradient normalization on the final target data as it continually adapts to the incoming batches of intermediate and target domains of AID-CC.}
    \label{fig:fluctuations-ucl}
  \end{subfigure}
  \caption{Continual models ConDA and UCL-GV with Swin backbone, showing increase in adaptation stability and final accuracy due to gradient normalization.}
  \label{fig:fluctuations_normalized_swin}
\end{figure*}

We can see from the results in Table \ref{tab:res_transformers1} that while gradient normalization may prevent optimal adaptation performance for the continual-SHOT model, adaptation stability increases for the two buffer-fed continual models ConDA and UCL-GV, and both models generally adapt better to the target domains at the end of the adaptation process. 
With the stability gained by gradient normalization, the models ConDA and UCL-GV can be considered comparable in adaptation performance. 
Accuracies of ConDA and UCL-GV on the target domain for continual adaptation for AID-CC, with Swin-B as the backbone are plotted in Figure \ref{fig:fluctuations_normalized_swin}, to inspect the increase of adaptation stability due to gradient normalization.

The results in Table \ref{tab:res_transformers2}, particularly those for UCM-CC, show the adverse effects of instability during adaptation. At the higher initial learning rate, and without stabilization by gradient normalization, the adaptation process may completely collapse. The impact is more severe for UCM-CC due to the higher degree of degradation of cloud cover in our datasets. 
However, when the gradients are normalized and the adaptation process is stabilized, the models show very promising performance. This clearly shows the effectiveness of gradient normalization and the resulting improvements over the original forms of the continual methods.
Consistent with the results for AID-CC and AID-SF, ConDA and UCL-GV with Swin-B backbone beat the other two backbone architectures for UCM-CC and UCM-SF.

Overall, the two continual DA models ConDA and UCL-GV outperform the standard SHOT model under continual setting. 
This shows the efficacy of selectively storing samples in a memory buffer and replaying these samples from earlier batches mixed with the samples from new incoming batch. 
Such memory replay helps in retaining knowledge gained from earlier batches, and results in a better domain adaptation to the target domain. 
Gradient normalization with smaller learning rates, despite preventing optimal adaptation in certain cases for smaller datasets, significantly increases adaptation stability, and prevents the models from potentially collapsing. 

In terms of the effectiveness of different backbones, Swin generally outperforms the CNN-based ResNet-50 model for continual DA on our benchmark evaluations. 
This can be attributed to its attention mechanism as well as the larger size of the model, which contribute to the increased ability of transformers to capture global feature representations at the lower layers compared to CNN-based architectures. 
Between the two transformer models evaluated, Swin consistently outperforms ViT, without significantly increasing the model size.
It is notable that 
ResNet-50 did better than ViT, even though it has a much lower number of parameters. 
This can be attributed to the weak inductive bias observed in ViT \cite{steiner2022how}, leading to increased overfitting on the source data. 
Raghu \textit{et al.} \cite{raghu2021vision} showed that the lower layer effective receptive fields for ViT are larger than those in ResNet and in the Swin transformer.
This results in weaker inductive bias, and therefore requires a large amount of data for effective training.
Swin transformer is a hierarchical transformer where self-attention is calculated within a local sliding window, that leads to stronger inductive bias and requires comparatively less data for training.
Due to its weaker inductive bias, ViT is not well-suited as a backbone for continual DA with single pass network updates on limited amount of data, compared to both ResNet-50 and Swin.

\section{Conclusion}

In this paper, we present four novel aerial domain adaptation benchmark datasets for developing domain adaptation models under gradually degrading weather conditions. 
We utilize two of the most common inclement weather conditions to synthesize gradually worsening weather on two popular aerial imagery datasets. 
We evaluate one standard DA model under the continual setting, and two state-of-the-art continual DA models on the new datasets, and provide benchmark results for future research. 
We specifically evaluate under the continual or test-time setting, to mimic scenarios where weather may gradually change, or degrade while the model is deployed in the real-world. 
We discover significant adaptation stability issues in buffer-fed continual DA models, and propose gradient normalization before model optimization as a tool to increase adaptation stability. 
Gradient normalization however slows the optimization process, and may prevent optimal continual adaptation on small datasets. 
We further assess the continual models with stronger attention-based transformer backbones and find that the Swin transformer performs the best, while ViT is not well-suited for continual DA.
We hope that this research will guide future directions not only in continual DA, but also in stable adaptation and training practices. 

\appendix    

\subsection*{Disclosures}
The authors declare no conflict of interest.

\subsection* {Code, Data, and Materials Availability} 
The datasets used are available at 
https://github.com/sadman-jahan/AID-UCM-DegradingWeather

\subsection* {Acknowledgments}
The authors would like to thank Dr. Erik Blasch for participating in the research conceptualization.
This research was partly supported by the Air Force Office of Scientific Research (AFOSR) under SBIR grant FA9550-22-P-0009 with Intelligent Fusion Technology, and AFOSR grant FA9550-20-1-0039. 
The authors would like to thank RIT Research Computing for making computing resources available for experimentation.


\bibliography{article}   
\bibliographystyle{spiejour}   




\listoffigures
\listoftables

\end{spacing}
\end{document}